\documentclass{article}
\usepackage{spconf,amssymb,amsmath,epsfig}
\usepackage{microtype}
\usepackage{graphicx}
\usepackage{multirow}
\usepackage{booktabs}
\usepackage{amssymb,amsmath,graphicx}
\usepackage{caption}
\usepackage{subcaption}
\usepackage[export]{adjustbox}
\usepackage{moredefs} 
\usepackage{color}

\newcommand{\red}[1]{{\color{red} \textbf{#1}}}

\defcommand{\vec}[1]{\mathbf{#1}} 

\def\vx{\mathbf{x}}
\def\vh{\mathbf{h}}
\def\vy{\mathbf{y}}

\def\eow{\texttt{<eow>}}

\title{No Need for a Lexicon? Evaluating the Value of the Pronunciation Lexica in End-to-End Models}
\name{{Tara N. Sainath, Rohit Prabhavalkar, Shankar Kumar, Seungji Lee, Anjuli Kannan, David Rybach}
  \secondlinename{Vlad Schogol, Patrick Nguyen, Bo Li, Yonghui Wu, Zhifeng Chen, Chung-Cheng Chiu}
\address{Google, Inc., New York, NY, USA \\
\fontsize{9}{9}\selectfont\ttfamily\upshape
\{tsainath,prabhavalkar,shankarkumar,leesj,anjuli,rybach\}@google.com \\
\fontsize{9}{9}\selectfont\ttfamily\upshape
\{vlads,drpng,boboli,yonghui,zhifengc,chungchengc\}@google.com}}

\begin{document}
\maketitle
\ninept
\begin{abstract}
For decades, context-dependent phonemes have been the dominant sub-word unit
for conventional acoustic modeling systems.
This status quo has begun to be challenged recently by end-to-end models which
seek to combine acoustic, pronunciation, and language model components into a
single neural network.
Such systems, which typically predict graphemes or words, simplify the
recognition process since they remove the need for a separate
expert-curated pronunciation lexicon to map from phoneme-based units to words.
However, there has been little previous work comparing phoneme-based versus
grapheme-based sub-word units in the end-to-end modeling framework, to determine
whether the gains from such approaches are primarily due to the new
probabilistic model, or from the joint learning of the various components with
grapheme-based units.

In this work, we conduct detailed experiments which are aimed at quantifying the
value of phoneme-based pronunciation lexica in the context of end-to-end models.
We examine phoneme-based end-to-end models, which are contrasted against grapheme-based ones on a large vocabulary
English Voice-search task, where we find that graphemes do indeed outperform
phonemes. We also compare grapheme and phoneme-based approaches on a
multi-dialect English task, which once again confirm the superiority of
graphemes, greatly simplifying the system for recognizing multiple dialects.
\end{abstract}

\section{Introduction \label{sec:introduction}}
Traditional automatic speech recognition (ASR) systems are comprised of an acoustic model (AM), a language model (LM) and a pronunciation model (PM), all of which are independently trained on different datasets. AMs take acoustic features and predict a set of sub-word units, typically context-dependent or context-independent phonemes. Next, a hand-designed lexicon (i.e., PM) maps a sequence of phonemes produced by the acoustic model to words. Finally, the LM assigns probabilities to word sequences.

There have been many attempts in the community to fold the AM and PM into one component \cite{McGraw2013,Lu2013}. This is particularly helpful when training multi-lingual systems \cite{Ney2003}, as a single AM+PM can be potentially used for all languages. A recent popular approach to jointly learn the AM+PM is to have a model directly predict graphemes. However, to date, most grapheme-based systems do not outperform phone-based systems \cite{Sung2003,Graves2009,Rao17a}.

More recently, there has been a growing popularity in end-to-end systems, which attempt to learn the AM, PM and LM together in one system. Most work to date has explored end-to-end models which predict either graphemes or wordpieces~\cite{Jan15,Chan15,Baidu,Rao17b}, which removes the need for a hand-designed lexicon as well. These end-to-end systems outperform systems which learn only an AM+PM jointly \cite{RohitSeq17}, though to date many of these systems still do not outperform conventional models trained with separate AM, PM and LMs.

This leads to the natural question: how do end-to-end models perform if we incorporate a separate PM and LM into the system? This question can be answered by training an end-to-end model to predict phonemes instead of graphemes. The output of the end-to-end model must then be combined with a separate PM and LM to decode the best hypotheses from the model. End-to-end phoneme models have been explored for a small-footprint keyword spotting task \cite{Ryan17}, where the authors found that models trained to predict phonemes were better than graphemes. However, this system produced a small number of keyword outputs, thus requiring a simple lexicon and no language model.
 In our previous work, we also demonstrated that phoneme-based end-to-end systems can be used to improve performance by rescoring lattices decoded from conventional ASR systems~\cite{RohitAnal17}. To our knowledge, the present work is the first to explore end-to-end systems trained with phonemes for a large vocabulary continuous speech recognition (LVCSR) task, \emph{where models are directly decoded in the first-pass}.

Our first set of experiments, conducted on a 12,500-hour English Voice Search task, explore the behavior of end-to-end systems trained to predict graphemes vs. phonemes. Our experiments show that the performance of grapheme systems is slightly better than phoneme systems. Since a benefit of end-to-end systems arises in systems trained for multiple dialects/languages, we extend our comparison towards a multi-dialect system trained on 6 different English dialects. Again, we find the grapheme system outperforms the phoneme system.

The rest of this paper is structured as follows. In Section 2 we describe training and decoding an end-to-end model with phonemes. The experimental setup is described in Section 3 and results are presented in Section 4. Finally, Section 5 concludes the paper and discusses future work.


\section{Incorporating Phonemes into \\ An End-To-End Model}\label{sec:cip}
\subsection{Components of Conventional ASR System}
Given an input sequence of frame-level features (e.g., log-mel-filterbank
energies), $\vx=\{x_1, x_2, \ldots, x_T\}$, and an output sequence of sub-word
units (e.g., graphemes, or phonemes), $\vy=\{y_1, y_2, \ldots y_N\}$, the goal of any speech
recognition system is to model the distribution over output sequences
conditioned on the input, $P(\vy|\vx)$. Typically, the process of finding the best set of recognized words in the network is represented as a composition of finite state transducers (FSTs) \cite{Mohri2008},
shown by Equation \ref{eq:transducers}.
\begin{equation}
D = C \circ L \circ G
\label{eq:transducers}
\end{equation}
An acoustic model is trained to map the input $\vx$ to a set of context-dependent phones.
With reference to Equation~\ref{eq:transducers}, a $C$ transducer maps context-dependent phones to context-independent phones (CIPs). The output of $C$ is composed with a pronunciation model, represented by an $L$ transducer. The $L$ transducer takes sequences of CIPs and maps them to words. Finally, the language model is represented by $G$, which
assigns probablities to sequences of words. A potential drawback with this approach is that the acoustic, pronunciation and language models are all trained separately. Furthermore, $L$ is manually curated and a challenging text normalization step is required to map between the verbalized and written representations.

\subsection{End-to-end models}

\begin{figure}[h!]
\centering
  \includegraphics[scale=0.35]{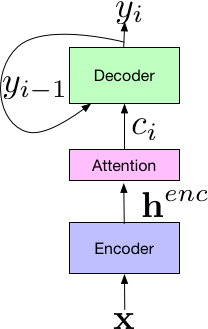}\\
  \caption{Components of the LAS end-to-end model.}
  \label{fig:las}
  \vspace{-0.1in}
\end{figure}

End-to-end models attempt to fold parts of the recognition process in Equation \ref{eq:transducers} into a single neural network. While there are many end-to-end models that have been explored, in this paper we will focus on attention-based models, such as Listen, Attend and Spell (LAS) \cite{Chan15}. This model consist of 3 modules as shown in Figure \ref{fig:las}.

The \emph{listener} module, also known as the encoder, takes the input features, $\vx$, and maps this to a higher order feature representation $\vh^{enc}$. We can think of the encoder as similar to a typical acoustic model. The output of the encoder is passsed to an \emph{attender}, which acts like an alignment mechanism, determining which encoder features in $\vh^{enc}$ should be attended to in order to predict the next output symbol, $y_i$. The output of the
attention module is passed to the \emph{speller} (i.e., decoder), which takes the
attention context, $c_i$, as well as an embedding of the previous prediction, $y_{i-1}$, in order to produce a probability distribution, $P(y_i|y_{i-1}, \ldots, y_0, \vx)$, over the current sub-word unit, $y_i$, given the previous units, $y_{i-1}, \ldots, y_0$, and input, $\vx$. We can think of the decoder as similar to a language model. The model also contains two additional symbols, namely a {\tt <sos>} token which is input to the decoder at time step $y_0$, indicating the start of sentence and an {\tt <eos>} to indicate end of sentence. The model is trained to minimize the cross-entropy loss on the training data.

\subsection{Grapheme Units}
Graphemes are a very common subword unit for end-to-end models.
In our work, the grapheme inventory includes the 26 lower-case letters a--z, the numerals $0$-–$9$, a label
representing \texttt{<space>}, and punctuation.

The decoding process involves finding the best grapheme sequence, $\vy^*$, under the model distribution, in other words:
\begin{equation}
  \vy^* = \arg \max_y p(\vy|\vx) = \arg \min_y -\log p(\vy|\vx)
  \label{eq:beam_search}
\end{equation}
Typically, decoding using an end-to-end model is performed using a beam search. At each step in the beam search, candidate hypotheses are formed by extending each hypothesis in the beam by one grapheme unit. These updated hypotheses are scored with the LAS model, and generally a small number (e.g., 8) of top-scoring candidates are kept to form a new beam for the next decoding step. The model stops decoding when the {\tt <eos>} symbol is predicted.

The prediction of graphemes allows us to remove the need for both the $C$ and $L$ transducers during decoding. This is because the graphemes that are produced by the beam search can simply be concatenated into words, with the predicted \texttt{<space>} token indicating word boundaries.

\subsection{Phoneme Units}
Instead of having the end-to-end model predict graphemes, the model can predict phonemes, but at the cost of requiring additional transducers during decoding which are not needed by the grapheme system.
In this work, we explore having the model predict context-independent phonemes (CIP), thus removing the need for a $C$ transducer. Following the small-footprint keyword spotting end-to-end work in \cite{Ryan17}, we train
our model to predict a set of 44 CI phonemes, as well as an extra \eow{} token, specifying the end of a word, analogous to the \texttt{<space>} token in graphemes (e.g., \texttt{the cat} $\to$ \texttt{d ax \eow{} k ae t \eow{}}).

Because of the homophone issue with phonemes (e.g., phoneme \texttt{ey} can map to the words `I' or `eye'), using a language model, $G$, is critically important. There are two ways we can incorporate $L$ and $G$ during decoding, mapping from a sequence of phonemes to a sequence of words: first, similar to graphemes, the output of the beam search can produce an n-best list of phonemes; each such phoneme sequence can be composed independently with $L$ and $G$ to get the n-best list of word sequences. This requires having an external LM weight $\lambda$ on $G$ in order to balance the scores coming from the end-to-end model relative to the scores from $L$ and $G$. We will refer to this strategy as \emph{N-best Combination}. As an alternative, we can bias each step of the beam search with $L\circ G$, similar to what was done in \cite{Chorowski17,Anjuli18} for graphemes. This strategy, which we will refer to as \emph{Beam-search Combination} is denoted by Equation \ref{eq:beam_search_cip}.
\begin{equation}
  \vy^* = \arg \min_y -\log p(\vy|\vx) - \lambda \log p_{LM} (\vy) - \eta \texttt{coverage}
  \label{eq:beam_search_cip}
\end{equation}

In this equation $ p(\vy | \vx)$ is the score from the LAS model, which is combined with a score coming from $L\circ G$ ($p_{LM}(\vx)$) weighted by an LM weight $\lambda$, and a \texttt{coverage} term to promote longer transcripts \cite{Chorowski17} and weighted by $\eta$. The benefit of this approach is that the $L$ and $G$ bias each step of the beam search rather than at the end, which is similar to our conventional models. However, one drawback is that as \cite{Chorowski17} indicates, the equation is a heuristic to combine independent models which can become quite challenging if the end-to-end model term $-\log p(\vy|\vx)$ becomes over-confident, in which case, the weight from the LM component will be ignored. We can also apply $L\circ G$ in both the beam-search and n-best, which will be explored as well.

\section{Experimental Details \label{sec:experiments}}
Our intial experiments are conducted on a $\sim$12,500 hour training set consisting of 15M US English utterances. The training utterances are anonymized and hand-transcribed, and are representative of Google's voice search traffic.
This data set is created by artificially corrupting clean utterances using a
room simulator, adding varying degrees of noise and reverberation such that the
overall SNR is between 0dB and 30dB, with an average SNR of 12dB \cite{Chanwoo17}. The noise sources are drawn from YouTube videos and daily life noisy environmental  recordings. We report results on a set of $\sim$14,800 anonymized, hand-transcribed Voice Search utterances extracted from Google traffic.

We also conduct experiments on 5 different English dialects, namely India (IN), Britain (GB), South Africa (ZA), Nigeria \& Ghana (NG) and Kenya
(KE). A single multi-dialect model is trained on these languages, totaling about 20M utterances (~27,500 hours). Noise is artificially added to the clean utterances using the same procedure for US English. We report results on a dialect-specific test set, which is around 10K utterances per test set. We refer the reader to \cite{Bo18} for more details about the experimental setup.

All English experiments use 80-dimensional log-mel features, computed with a 25-ms window and shifted every 10ms. Similar to~\cite{Hasim15, Golan16}, at the current frame, $t$, these features are stacked, with 3 frames to the left (for US English) and 7 frames for multi-dialect, and downsampled to a 30ms frame rate. The encoder network architecture consists of 5 unidirectional long short-term memory~\cite{HochreiterSchmidhuber97} (LSTM) layers, with the size specified in the results section.  Additive attention is used for all experiments \cite{Bahdanau14}. The decoder network is a 2 layer LSTM with 1,024 hidden units per layer. The grapheme systems use 74 symbols while the phoneme systems use 45 CIP for US English, and a unified set of 50 CIPs for multi-dialect.

All neural networks are trained with the cross-entropy criterion, using
asynchronous stochastic gradient descent (ASGD) optimization~\cite{Dean12} with Adam~\cite{KingmaBa15} and are trained using TensorFlow~\cite{AbadiAgarwalBarhamEtAl15}.

\section{Results \label{sec:results}}
\subsection{Tuning CIP Models}
Our first set of experiments explore what parameters are important for decoding an end-to-end model trained with CIP.

\subsubsection{Tuning LM Weight}
\begin{figure} [h!]
\centering
  \includegraphics[scale=0.35]{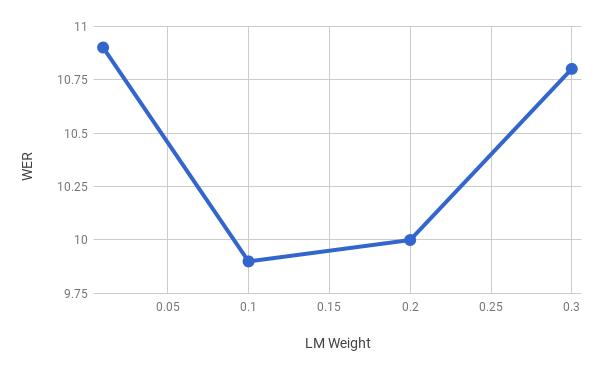}\\
  \caption{WER as a function of LM weight.}
  \label{fig:werlmw}
\end{figure}

First, we explore the behavior of the language model weight (LMW), $\lambda$, when $L \circ G$ is incorporated using \emph{N-best Combination} following beam search.
Figure \ref{fig:werlmw} shows the WER as a function of the LMW. The figure indicates that WER is heavily affected by the choice of LMW, which seems to be best around $0.1$. This also indicates a drawback of using phonemes, namely an external weight needs to be tuned to balance the scores coming from the end-to-end model relative to $L \circ G$. This can be a drawback if the end-to-end model is overconfident and produces a high probability, thus de-emphazising the score from the language model component.

\subsubsection{Incorporating End-of-Word Symbol}
Next, we explore different ways of using the \eow{} symbol during decoding. In \cite{Ryan17}, the \eow{} symbol was required during decoding and was shown to help in identifying the spacing between words. In particular, since models were decoded without a separate LM, requiring an \eow{} symbol between words was found to be critical to minimize false positives (e.g., to avoid false triggering on the phrase \texttt{America}, for the keyword \texttt{Erica}). However, in a LVCSR task like Voice Search where we use a separate PM and LM for $L$ and $G$, requiring \eow{} might cause the model to make errors and predict incorrect words if \eow{} is not correctly predicted. Table \ref{table:phoneme_eow} shows that it is better to make \eow{} optional rather than required. Note for these experiments, $L\circ G$ is incorporated using \emph{N-best Combination}.

\begin{table} [h!]
\centering
\begin{tabular}{|c|c|} \hline
System & WER \\ \hline
LAS unid, required \eow{}  & 10.2 \\ \hline
LAS unid, optional \eow{}  & 9.7 \\ \hline
\end{tabular}
\vspace{-0.1 in}
	\caption{WER Phoneme \eow{} analysis. Because it is hard to predict \eow{}, it is better to make it optional.}
\vspace{-0.1 in}
\label{table:phoneme_eow}
\end{table}

\subsubsection{Where to use $L \circ G$}
Finally, we study where to apply $L\circ G$, specifically if it should be applied either during the beam search \emph{Beam-search Combination}, following the beam search \emph{N-best Combination}, or in both places. Applying $L \circ G$ in both places requires tuning two separate LMW weights, for the $G$ during and after the beam search.

\begin{figure} [h!]
\centering
  \includegraphics[scale=0.35]{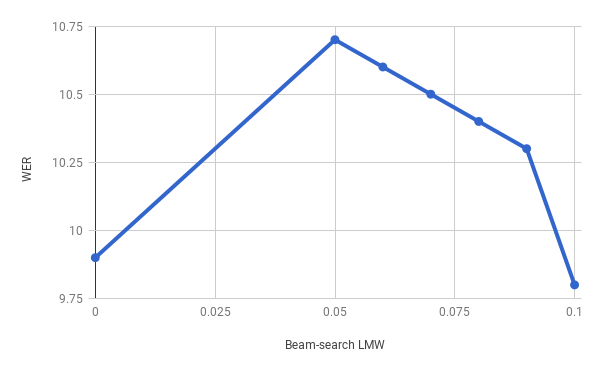}\\
  \caption{WER as a function of beam-search LM Weight. }
  \label{fig:werlmwbs}
\end{figure}

Figure \ref{fig:werlmwbs} shows the WER of the final system as the beam search LMW is increased from 0.0 to 0.1. For illustrative purposes, we set the weight of the first and second LM to sum to $0.1$, though more extensive sweeping of both LMWs did not improve performance further. The figure shows that a slight improvement is obtained with \emph{N-best Combination} (9.7, LMW$=0.0$) compared to \emph{Beam-search Combination} (9.8, LMW=$0.1$). This illustrates that the decoder of the LAS model is strong enough to learn the correct phone sequence, and thus \emph{N-best Combination} is sufficient enough to yield reasonable results. The rest of the results in this paper are reported with \emph{N-best Combination}.



\subsection{Phoneme vs. Grapheme Comparison, Model Architecture}

Having established a good recipe for training with CIP, we now compare the performance of phoneme and grapheme systems for different LAS architectures, namely a single head unidirectional LAS system and a multi-head unidirectional LAS system. We are specifically interested in multi-head (MHA) for LAS, as MHA has been shown to give state-of-the-art performance for LAS grapheme systems \cite{CC18}. MHA  allow the end-to-end model to jointly attend to information at different positions in the encoder space with multiple attention heads. The single head model is a 5x1024 encoder with 1 attention head and 2x1024 decoder. The MHA model is a 5x1400 encoder with 4 attention heads, followed by a 2x1024 decoder.


Table \ref{table:arch_compare} indicates that for single-head attention, both phoneme and grapheme systems have similar performance. However, for MHA the phoneme system lags behind the grapheme system. One hypothesis we have is as the end-to-end model, including the encoder and decoder, gets stronger, from single to multi-head attention, having a model which jointly integrates the AM, PM and LM (i.e., training with graphemes) is better than separate integration (i.e., training with phonemes).

\begin{table} [h!]
\centering
\begin{tabular}{|c||c|c|} \hline
Model & Phoneme & Grapheme \\ \hline
unid LAS & 9.7  & 9.8  \\ \hline
MHA LAS & 8.6 (1.4/1.8/5.4) & 8.0 (1.1/1.3/5.6)  \\ \hline
\end{tabular}
\vspace{-0.1 in}
\caption{CIP vs. Graphemes Across Model Architectures. The (del/ins/sub) is indicated in parenthesis.}
\vspace{-0.1 in}
\label{table:arch_compare}
\end{table}
To understand the errors made by phonemes and graphemes, we pulled a few representative examples. Table \ref{table:grapheme_wins} shows examples of where the grapheme system wins over the phoneme system. The first example in the table indicates that the phoneme system has slightly higher deletions, likely because of the incorporation of the external $L$ and $G$ and the need to tune an LMW. This is also confirmed quantitiatively by looking at the deletions, insertions, and substitutions in Table \ref{table:arch_compare}. In addition, because of the hand-designed lexicon $L$, the second example shows that phoneme system does not do as well with text normalization. Finally, the grapheme system benefits from making a joint decision for disambiguating homophones while the chained phoneme system does not, as shown in the third example.

\begin{table} [h!]
\centering
\begin{tabular}{|c|c|} \hline
Grapheme &  Phoneme \\ \hline
let me see a clown & Let me see \\ \hline
How old is 50 cents & How old is \red{\$0.50} \\ \hline
Easy Metallica songs to & \red{AZ} Metallica songs to  \\
play on the guitar &  play on the guitar \\ \hline
\end{tabular}
\vspace{-0.1 in}
\caption{Grapheme Wins. Phoneme errors indicated in \red{red}.}
\vspace{-0.1 in}
\label{table:grapheme_wins}
\end{table}

In contrast, Table \ref{table:phoneme_wins} gives examples where the phoneme system wins over the grapheme system. The phoneme system wins on proper nouns and rare words, aided by the hand designed lexicon $L$ and the LM $G$, which is trained on a billion word text-only corpora.

\begin{table} [h!]
\centering
\begin{tabular}{|c|c|} \hline
Grapheme &  Phoneme \\ \hline
Albert Einstein versus & Albert Einstein vs. \\
\red{Singapore} & Stephen Hawking \\ \hline
Head Start on & Head Start Ronkonkoma \\
\red{Concord} New York & New York \\ \hline
Charles \red{Lindberg} in Paris & Charles Lindbergh in Paris \\ \hline
\end{tabular}
\vspace{-0.1 in}
\caption{Phoneme Wins. Grapheme errors indicated in \red{red}.}
\vspace{-0.1 in}
\label{table:phoneme_wins}
\end{table}

\subsection{Comparison for Multi-dialect}

In this section, we compare the performance of phones vs. graphemes for a multi-dialect English system. Both systems use a 5x1024 encoder with single-head attention, followed by a 2x1024 decoder. For the phoneme systems, we use a unified phone set and a unified $L$, but a language-specific $G$ for each language. The results are shown in Table \ref{table:enx_cip}. The table shows that across the board the phoneme system is worse than the grapheme system. 

Table \ref{table:grapheme_wins_enx} gives a few examples of where the phoneme system makes errors compared to the grapheme system. In addition to text norm and deletion errors like in English, the multi-dialect phone system also has many pronunciation errors. The table illustrates this the disadvantage of having a hand-designed lexicon. Overall, a grapheme end-to-end model provides a much simpler and more effective strategy for multi-dialect ASR.

\begin{table} [h!]
\centering
\begin{tabular}{|c||c|c|c|c|c|} \hline
Dialect & IN & GB & ZA &  NG & KE \\ \hline
grapheme & 18.4 & 14.1 & 13.8 & 34.5 & 19.9 \\ \hline
phoneme & 31.6 & 18.1 & 18.6 & 39.0 & 24.8 \\ \hline
\end{tabular}
\vspace{-0.1 in}
\caption{WER of CIP vs. Graphemes For Mulit-dialect}
\vspace{-0.1 in}
\label{table:enx_cip}
\end{table}

\begin{table} [h!]
\centering
\begin{tabular}{|c|c|} \hline
Grapheme &  Phoneme \\ \hline
Chris Moyles bake off & Chris \red{Miles} bake off  \\ \hline
Ukip Sussex candidates & \red{You} kip Sussex candidates \\ \hline
What does Allison mean & What does \red{Alison} mean \\ \hline
My name is Reese & My name is \red{Rhys} \\ \hline
\end{tabular}
\vspace{-0.1 in}
\caption{Grapheme Wins for Multi-dialect. Phoneme errors indicated in \red{red}.}
\vspace{-0.1 in}
\label{table:grapheme_wins_enx}
\end{table}

\section{Conclusions \label{sec:conclusions}}

In this paper, we examined the value of a phoneme-based pronunciation lexica in the context of end-to-end models. Specifically, we compared using phone vs. grapheme systems with an end-to-end attention-based model. We found that for both US English and multi-dialect English, the grapheme systems were superior to the phone systems. Error analysis shows that the grapheme systems lose on proper nouns and rare words, where the hand-designed lexica help. Future work will look at combining the strengths of both of these units into one system.

\section{Acknowledgements}
The authors would like to thank Eugene Weinstein and Michiel Bacchiani for useful discussions. In addition, thanks to Alyson Pitts, Jeremy O'Brien, Shayna Lurya and Evan Crewe for help with the multi-dialect experiments.
\bibliographystyle{IEEEbib}
\bibliography{main}
\end{document}